\documentclass[sigconf]{acmart}

\usepackage{booktabs} 
\usepackage{graphicx}
\usepackage{amsmath,amssymb} 
\usepackage[utf8]{inputenc}
\usepackage{enumitem}
\usepackage{subcaption}
\usepackage[sort&compress,capitalize,noabbrev,nameinlink]{cleveref}
\creflabelformat{equation}{#2#1#3}
\setcopyright{rightsretained}

\acmDOI{}

\acmISBN{}

\acmConference[ACM Computer Science in Cars Symposium]{ACM Computer Science in Cars Symposium}{2018}{Munich, Germany}
\acmYear{2018}

\acmSubmissionID{33}

\begin{document}
\title{
Automated Scene Flow Data Generation \\ 
for Training and Verification
}
\subtitle{Extended Abstract}

\author{Oliver Wasenmüller}
\author{René Schuster}
\author{Didier Stricker}
\affiliation{
  \institution{DFKI}
}
\email{firstname.lastname@dfki.de}

\author{Karl Leiss}
\author{J\"urger Pfister}
\author{Oleksandra Ganus}
\affiliation{
  \institution{Bit-TS}
}
\email{firstname.lastname@bit-ts.de}

\author{Julian Tatsch}
\author{Artem Savkin}
\author{Nikolas Brasch}
\affiliation{
  \institution{BMW}
}
\email{firstname.lastname@bmw.de}

\renewcommand{\shortauthors}{Wasenm\"uller et al.}

\begin{teaserfigure}
	\begin{center}
		\begin{subfigure}[c]{0.32\textwidth}
			\centering
			\includegraphics[width=1\textwidth]{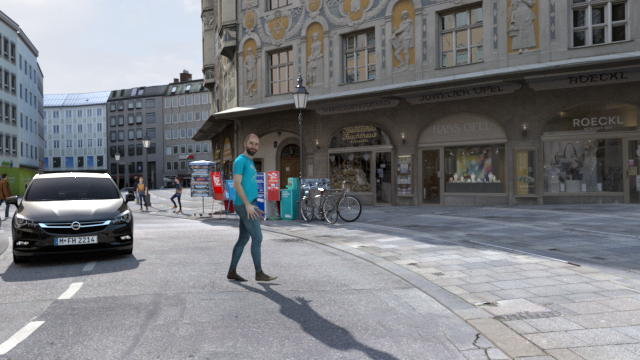}
			\subcaption{Image}
		\end{subfigure}
		\begin{subfigure}[c]{0.32\textwidth}
			\centering
			\includegraphics[width=1\textwidth]{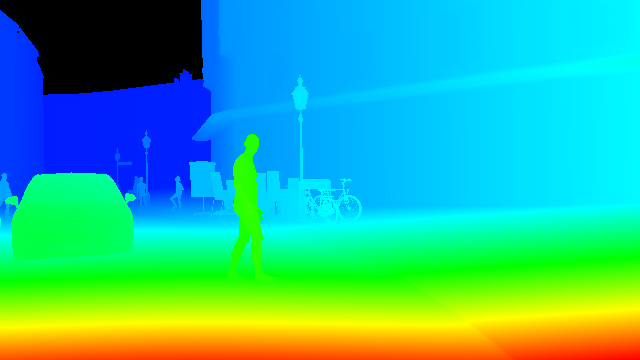}
			\subcaption{Depth}
		\end{subfigure}
		\begin{subfigure}[c]{0.32\textwidth}
			\centering
			\includegraphics[width=1\textwidth]{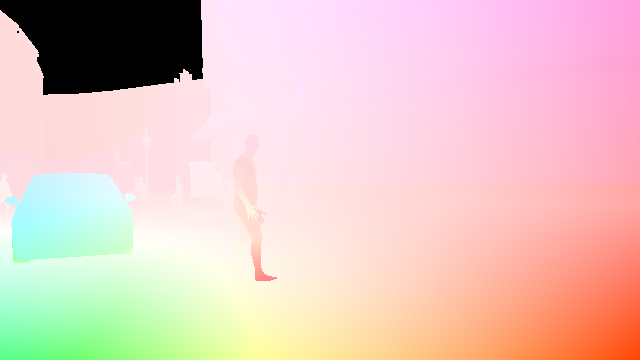}
			\subcaption{Flow}
		\end{subfigure}
		\caption{
		We present a technology to automate the creation of synthetic images with dense scene flow ground truth.
		}
		\label{fig:teaser1}
	\end{center}
\end{teaserfigure}

\begin{abstract}
Scene flow describes the 3D position as well as the 3D motion of each pixel in an image.
Such algorithms are the basis for many state-of-the-art autonomous or automated driving functions.
For verification and training large amounts of ground truth data is required, which is not available for real data.
In this paper, we demonstrate a technology to create synthetic data with dense and precise scene flow ground truth.
\end{abstract}

%
%



\settopmatter{printacmref=false,printccs=false}

\maketitle


\section{Introduction}
\label{sec:intro}

In the rapid development of autonomous driving functions and advanced driver assistance systems (ADAS), the motion estimation of all objects around the vehicle plays an essential role.
Nowadays new cars are equipped with a multitude of sensors such as cameras, radar and ToF \cite{yoshida2017time}.
In order to determine the 3D motion of surrounding vehicles, pedestrians or unknown objects, so-called scene flow algorithms are utilized.
Scene flow algorithms compute for each pixel of a stereo camera the 3D position as well as the 3D motion.
This information can be used as an input for different driving functionalities.
Many state-of-the-art algorithms for computing scene flow geometrically have been presented in the Computer Vision literature \cite{behl2017bounding,menze2015object,lv2016CSF,ren2017cascaded,schuster2018sceneflowfields,vogel2015PRSM}.
For their verification and evaluation dense ground truth data is required.
Recently, there is a trend towards machine learning based flow estimation \cite{guney2016deep,ilg2017flownet,meister2018unflow,thakur2018sceneednet}.
These algorithms require even larger amounts of representative data for training and validation.
While manual ground truth labeling on real data might be somewhat achievable for high-level annotations (e.g. 2D/3D bounding boxes, lane markings, etc.),  precise labeling on pixel level is technically impossible.
Thus, we demonstrate in this paper a technology to create synthetic data with dense ground truth.
After discussing state-of-the-art datasets in Section \ref{sec:related}, we propose in Section \ref{sec:scene} an approach the render scene flow.
In Section \ref{sec:data} a technology for automated scene creation is presented followed by the data verification in Section \ref{sec:veri}.
%


\section{Related Work}
\label{sec:related}

As mentioned before, manual data labeling for scene flow is not possible.
Even an expert can not determine for each pixel and with high precision its corresponding pixel in other images.
There is also no hardware device, which can measure scene flow directly.
Some datasets, like the famous KITTI \cite{geiger2012kitti,menze2015object}, use a lidar with many scan lines in combination with a high-precision localization system.
With this configuration it is possible to compute scene flow for static scene content.
However, this flow is sparse and invalid for dynamic scene content.
Thus, we rely on synthetic data, where dense ground truth can be generated for any scene content.
In state-of-the-art synthetic datasets, such as Sintel \cite{butler2012sintel}, virtualKITTI \cite{gaidon2016vkitti} or P4B \cite{richter2017playing}, scene flow is not yet included.

\begin{figure*}[t]
	\begin{center}
		\begin{subfigure}[c]{0.32\textwidth}
			\centering
			\includegraphics[width=1\textwidth]{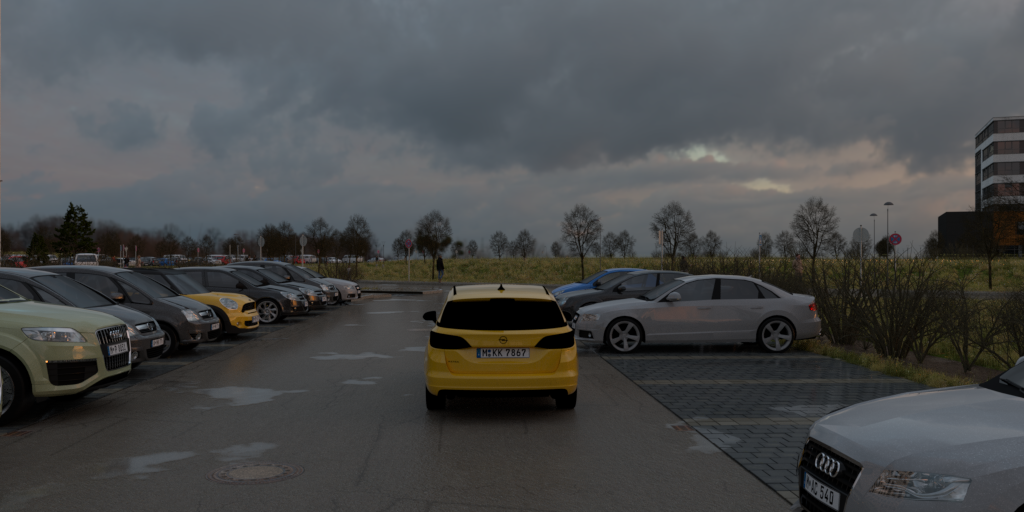}
			\subcaption{Image}
		\end{subfigure}
		\begin{subfigure}[c]{0.32\textwidth}
			\centering
			\includegraphics[width=1\textwidth]{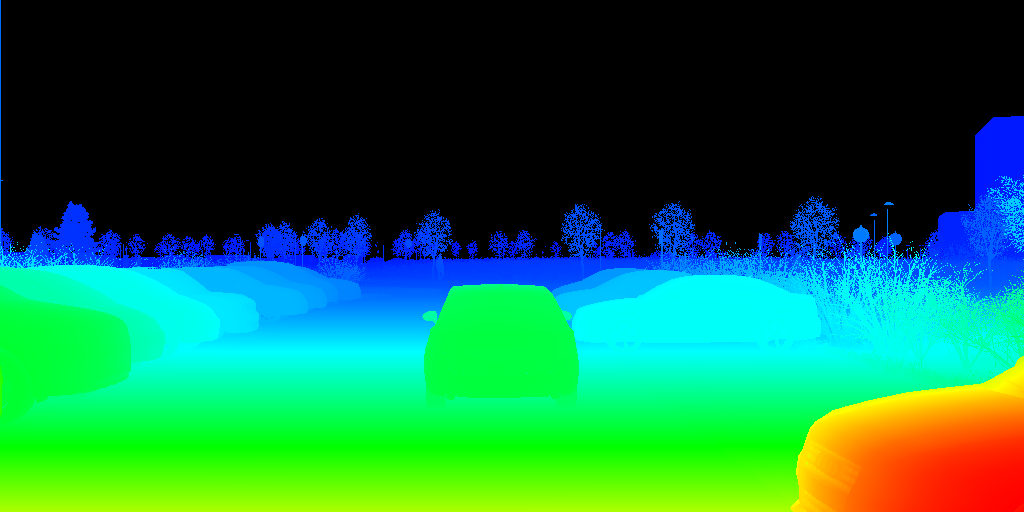}
			\subcaption{Depth}
		\end{subfigure}
		\begin{subfigure}[c]{0.32\textwidth}
			\centering
			\includegraphics[width=1\textwidth]{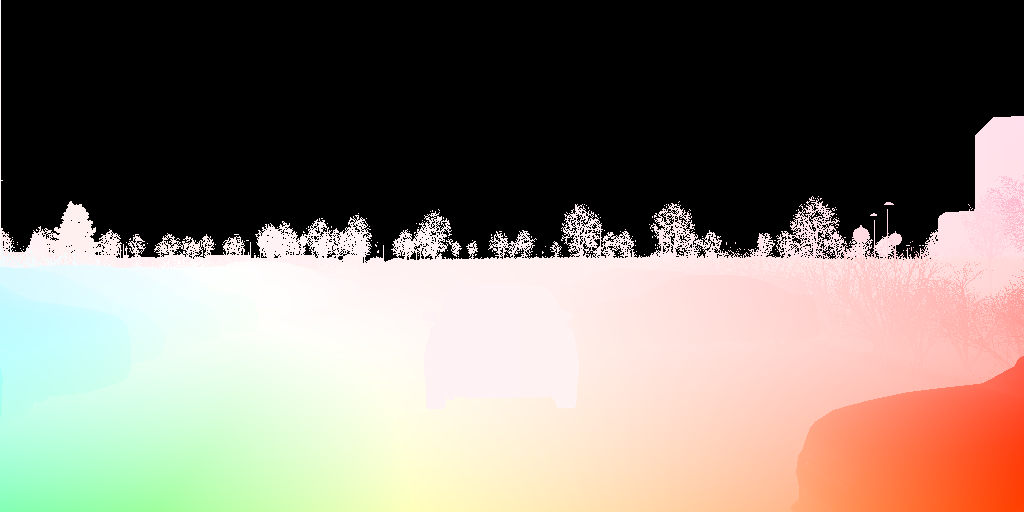}
			\subcaption{Flow}
		\end{subfigure}
		\caption{
		Scene flow for (a) an image is illustrated as (b) depth image for the 3D positions and (c) flow image for the 3D motion.
		}
		\label{fig:teaser1}
	\end{center}
\end{figure*}


\section{Scene Flow Rendering}
\label{sec:scene}

Scene flow represents the 3D position as well as the 3D motion of each pixel in a reference frame of a stereo camera.
In a synthetic scene it is straight forward to estimate the 3D position of each pixel with a simple back-projection.
The more tricky part is the motion estimation, since the camera motion as well as intrinsic motions and deformations of the scene need to be considered.
First, we determine the 3D position of each pixel by projecting a ray through each pixel in the 2D image plane and check for the surface point hit of the ray.
Besides the 3D coordinates of this hit, we store the corresponding triangle and its vertices.
This is necessary to track the 3D coordinates also under scene motion and deformation.
In case the scene contains intrinsic motion and/or deformation, the position of the triangle vertices will change in the next time step.
However, since we stored their identifier, we can determine the position of the hit also after intrinsic motion and/or deformation.
As long as the same ratio of distances to the vertices is maintained, the hit on a triangle can be determined precisely even after deformation.
In a second step, the camera ego-motion needs to be considered.
For that purpose, the hits after consideration of intrinsic motion and/or deformation are multiplied with the camera ego-motion.
This gives us the final position of each pixel at the next time step with respect to the reference frame.
The 3D motion of the scene flow are the 3D vectors between the starting hit and the computed hit.
With this technology the scene flow can be determined even for pixels moving out of the image.
%


\section{Scene Creation}
\label{sec:data}

The production of synthetic training and validation data has three main aspects to achieve for a long term impact:
First the physical based sensor impression needs to be met.  
Second, the content and production of data must be verifiable and fulfill a quality assurance process. 
Third, the scalability of data production is essential for training, test coverage and validation; especially for machine learning algorithms.
Today it is not proven which aspects of the real world are relevant for machine learning. 
Therefore, it is necessary to model a virtual sensor very close to the real sensor properties. 
Traditional content production workflows from game and film industry are based on thinking in whole scenes and using hand modeled content from artists. 
The artist can easily add additional meta-information into the scene so that labeling for a specific use case can be done automatically \cite{mayer2018makes}. 
Unfortunately, in this approach the quality is highly depending on the artist and provides very low flexibility to adopt data to new requirements regarding sensor models as well as scene variation. 
Our technology combines procedural and AI based generation of scenes on top of accurate maps including their semantic. 
Due to standardization, traffic signs and traffic lights are created according to their specifications. 
%
%
To organize assets, such as traffic signs, trees, vehicles, houses, etc., to reuse and automatically generate scene content in an efficient way, a database is built upon unique identifiers and a flexible data model.
The automation in the scenario building process is established by the categorization of assets in the database in a modular clustering of geometry, materials and meta data. 
By linking meta data dynamically together along with the street semantic a scene and its variations are generated. 
Thereby boundary conditions such as weather, local and global illumination, traffic conditions, trajectories and injection of noisy data along their corner cases can be set as parameters. 
The pipeline scales with the number of hardware threads and is thus highly parallelized. 
Same scalability applies for a later rendering step where the sensor and ground truth outputs are processed.
%
%


\section{Data Verification}
\label{sec:veri}

Although synthetic data allows to generate dense ground truth for scene flow, it is not guaranteed that the process is flawless. 
In order to validate the correctness of our data, we use automated sanity checks to verify its consistency. 
These checks include a round-trip check, forward-backward consistency, and ego-motion consistency. 
For a round-trip \cite{wasenmuller2014correspondence}, we use the generated ground truth for motion and geometry to traverse two stereo image pairs in a cyclic order. 
From reference, to next time step, to stereo camera, back to the previous time step, and finally back to the reference camera. 
Up to subpixel errors through rounding, a round trip should always reach its starting pixel if the ground truth is consistent. 
For forward-backward consistency checks, the additional generation of scene flow in inverted temporal order is required, i.e motion from a reference time to the previously generated image. 
The corresponding backward motion of the target pixel of the forward motion should be the inverse. 
Having full control over the rendering process also allows to compute the motion of the virtual observer. 
By using  ego-motion and  ground truth depth, the correctness of the scene flow for static parts of the scene can be verified. 
These three tests help to generate accurate and correct data for training and evaluation.
%


\section{Conclusion}
\label{sec:conc}

In this paper, we presented a technology to automate the creation of synthetic scene content for dense accurate scene flow.
Such data is required to train and verify different algorithms for driving functions in the context of autonomous driving and ADAS.
%


\bibliographystyle{ACM-Reference-Format}
\bibliography{bibcscs}

\end{document}